\documentclass[10pt, a4paper]{article}

\usepackage[final]{lrec2026} 
\usepackage [utf8]{inputenc}
\usepackage{multirow}
\usepackage{graphicx}
\usepackage{booktabs}
\usepackage{fvextra}
\usepackage{tcolorbox}
\usepackage{xurl}
\usepackage{amssymb}

\title{GaelEval: Benchmarking LLM Performance for Scottish Gaelic}

\name{\parbox{\textwidth}{\centering 
    Peter Devine,$^1$ William Lamb,$^1$ Beatrice Alex,$^1$ Ignatius Ezeani,$^2$ 
    Dawn Knight,$^3$ Mícheál J. Ó Meachair,$^4$ Paul Rayson,$^2$ Martin Wynne$^5$
}}

\address{$^1$University of Edinburgh, $^2$Lancaster University, $^3$University of Cardiff, \\
         $^4$Dublin City University, $^5$University of Oxford \\
         \{pdevine2, w.lamb, b.alex\}@ed.ac.uk, \{i.ezeani, p.rayson\}@lancaster.ac.uk, \\
         knightd5@cardiff.ac.uk, micheal.omeachair@dcu.ie, martin.wynne@ling-phil.ox.ac.uk}

\abstract{
Multilingual large language models (LLMs) often exhibit emergent `shadow' capabilities in languages without official support, yet their performance on these languages remains uneven and under-measured. This is particularly acute for morphosyntactically rich minority languages such as Scottish Gaelic, where translation benchmarks fail to capture structural competence. We introduce \textbf{GaelEval}, the first multi-dimensional benchmark for Gaelic, comprising: (i) an expert-authored morphosyntactic MCQA task; (ii) a culturally grounded translation benchmark and (iii) a large-scale cultural knowledge Q\&A task. Evaluating 19 LLMs against a fluent-speaker human baseline ($n=30$), we find that Gemini~3~Pro~Preview achieves 83.3\% accuracy on the linguistic task, surpassing the human baseline ($78.1$\%). Proprietary models consistently outperform open-weight systems, and in-language (Gaelic) prompting yields a small but stable advantage (+$2.4\%$). On the cultural task, leading models exceed $90$\% accuracy, though most systems perform worse under Gaelic prompting and absolute scores are inflated relative to the manual benchmark. Overall, GaelEval reveals that frontier models achieve above-human performance on several dimensions of Gaelic grammar, demonstrates the effect of Gaelic prompting and shows a consistent performance gap favouring proprietary over open-weight models.
}

\begin{document}
\maketitleabstract

\section{Introduction}
Although most large language models (LLMs) officially support a small fraction of the approximately 7,000 human languages spoken worldwide, they display emergent `shadow' capacities in many more. For instance, \mbox{OpenAI} advertises support for 59 languages in ChatGPT,\footnote{\url{https://help.openai.com/en/articles/8357869-how-to-change-your-language-setting-in-chatgpt}. Accessed 22 Feb 2026.} none of which belong to the Celtic family (e.g.~Irish, Welsh and Scottish Gaelic). Despite this, the system processes and generates text in every Celtic language. The distinction between official and \textit{de facto} support raises a methodological challenge: as coverage expands and model varieties diversify, establishing robust evaluation frameworks becomes crucial for both official languages and the minority languages they nevertheless represent.

The current evaluation landscape is markedly skewed. High-resource languages like English benefit from a self-reinforcing ecosystem of training corpora and mature benchmarks. In contrast, low-resource languages suffer from sparse training data \cite{joshi-etal-2020-state} and little or no evaluation resources \cite{romanou2024include}. Even where benchmarks exist, they rarely include human baselines \cite{assadi2025humemeasuringhumanmodelperformance}, making it impossible to ascertain whether a model's output follows a given community's sociolinguistic norms or not. Furthermore, the English-dominance of these models risks processing the world’s cultural-linguistic mosaic through an Anglocentric lens. Without objective measurement, academics and language communities alike cannot determine if an LLM should be explored or eschewed.  

Scottish Gaelic (`Gaelic') epitomises these challenges. Ranking 104th in Common Crawl accessibility,\footnote{\url{https://commoncrawl.github.io/cc-crawl-statistics/plots/languages}. Accessed 21 Feb 2026.} Gaelic occupies the digital margins, yet it possesses a rich morphosyntax that defies the only benchmarks available for it: the surface-level translation based FLORES-200 \citep{goyal2022flores} and BritEval \cite{britllm2026}. These benchmarks do not capture whether a model is truly `Gaelic-conversant' or merely performing a high-dimensional translation of English concepts. 

To address this gap, we present \textbf{GaelEval}: a targeted, multi-dimensional evaluation suite that moves beyond surface equivalence toward deeper morphosyntactic and culturally grounded competence. Our framework includes three distinct tasks:
\begin{enumerate}
    \item \textbf{Linguistic Competence}: A multiple-choice question answering (MCQA) task comprising 120-questions and probing fine-grained grammatical and idiomatic usage.

    \item \textbf{Translation}: A rigorous assessment using BLEU and \textsc{chrF} metrics against hand-translated gold labels.

    \item \textbf{Cultural Understanding}: A culturally grounded Q\&A task (1,087 questions) derived from pedagogical content produced by fluent speakers.
\end{enumerate}

We evaluate 19 contemporary LLMs (14 proprietary; 5 open-weight), providing the first systematic comparison of LLM performance for Scottish Gaelic. Gemini~3~Pro~Preview leads overall and surpasses the fluent-speaker baseline on the linguistic competence task.

Our principal contributions are:

\begin{itemize}
    \item \textbf{GaelEval}, the first multi-dimensional benchmark for Scottish Gaelic, spanning morphosyntax, translation and culturally grounded knowledge;
    \item the first human baseline for Gaelic LLM evaluation ($n=30$);
    \item evidence that Gemini~3~Pro~Preview exceeds the fluent-speaker mean on a controlled morphosyntactic task;
    \item a consistent aggregate advantage for in-language (Gaelic) prompting for the morphosyntactic task; and
    \item quantification of the performance gap between frontier proprietary and open-weight models in a minority-language setting.
\end{itemize}

In what follows, we review related work (§\ref{sec:related_work}), describe our design and evaluation methodology (§\ref{sec:benchmark_design}), present empirical results (§\ref{sec:results}) and conclude, with proposed directions for future work (§\ref{sec:conclusion}).

\section{Related Work}
\label{sec:related_work}
\paragraph{Multilingual LLM Evaluation Frameworks}
Large-scale multilingual benchmarks are central to evaluating LLM capabilities across languages. MMLU \citep{hendrycks2020measuring} introduced a widely used multiple-choice framework for knowledge-intensive reasoning in English. Global MMLU \citep{singh2025global} extended this paradigm cross-lingually, largely via translation of English-source materials. FLORES-200 \citep{goyal2022flores, nllb-24} expanded coverage to 200+ languages, including Gaelic, but evaluates only machine translation (MT). BritEval \cite{britllm2026} consists of 3 major English benchmarks translated into 4 languages from Britain and Ireland, including Gaelic. XTREME-UP \citep{ruder2023xtreme} incorporates additional low-resource tasks (e.g., transliteration, OCR), while INCLUDE \citep{romanou2024include} departs from translation-based design by constructing question answering benchmarks from native regional exam materials.

Many multilingual benchmarks rely heavily on translation or adaptation from English-centric datasets. While valuable, this approach under-represents language-specific morphosyntax, culturally grounded knowledge, and idiomatic usages that resist direct translation (e.g.\ that the colour of grass in Gaelic is \textit{gorm} `lit.~blue', not \textit{uaine} `lit.~green'). For morphologically rich languages such as Gaelic, translation-based evaluation also is unlikely to capture fine-grained inflectional contrasts or edge cases that distinguish structural competence from superficial word recognition. To our knowledge, beyond BritEval, FLORES-200 and related benchmarks \cite[e.g., SIB-200;][]{adelani2024sib}, no large-scale Gaelic evaluation suite exists.

\paragraph{Low-Resource and Morphologically Rich Language Evaluation}

Recent work addresses the challenges of evaluating LLMs on low-resource and morphologically complex languages, including tokenisation and pattern extrapolation \cite{xia2025evaluating}. IndicGenBench \cite{singh2024indicgenbench} covers 29 Indic languages using human-curated parallel data; AfriQA \cite{ogundepo2023afriqa} introduces question answering for African languages; and TurkBench \cite{toraman2026turkbench} evaluates Turkish across 21 subtasks. \citet{xia2025evaluating} further propose a cross-lingual benchmark spanning Cantonese, Japanese and Turkish, combining human evaluation with automated metrics across diverse tasks. Irish-BLiMP evaluates LLMs on Irish linguistic knowledge using 1020 minimal pairs and provides a human baseline \cite{mcgiff2025irish}. Collectively, these efforts highlight the need to evaluate LLMs on morphologically rich, low-resource languages. We extend this line of work by directly assessing model competence in Gaelic morphosyntax and non-compositional usage.

\paragraph{Culturally and Linguistically Informed Evaluation}

A growing literature argues that linguistic competence cannot be evaluated independently from cultural knowledge. \citet{tao2024cultural} document systematic bias toward English-speaking contexts in ostensibly multilingual LLMs. Relatedly, multilingual models have been shown to process non-English inputs through English-dominant representational pathways \citep{papadimitriou-etal-2023-multilingual, Wendler2024}, raising concerns about whether these systems encode language-specific structures or just rely on Anglocentric priors.

Recent benchmarks increasingly integrate cultural and linguistic evaluation. For example, ProverbEval \citep{azime2025proverbeval} assesses Ethiopian languages (and English) through proverb interpretation, requiring both morphosyntactic competence and culturally grounded reasoning. Knowledge-grounded benchmarks similarly test community-specific factual knowledge in domains such as food, holidays and social practices \citep{myung2025blendbenchmarkllmseveryday}. Importantly, \citet{myung2025blendbenchmarkllmseveryday} show that in-language prompting benefits medium- and high-resource languages, whereas low-resource languages often perform better under English prompting. This resource-sensitive pattern motivates our evaluation under both English and Gaelic prompt conditions to test whether similar asymmetries arise for Gaelic.

\paragraph{LLM-Assisted Benchmark Construction}
LLM-assisted benchmark generation offers a practical solution when human-curated datasets are scarce~\cite{perez2023discovering, anwar2026mcqs}. Prior work has used LLMs to extract cultural knowledge from large corpora such as C4~\cite{nguyen2023extracting} and TikTok~\cite{shi2024culturebank}, derive culturally grounded Q\&A from web scrapes~\cite{wang2024craft} and Wikipedia~\cite{fung2024no}, and generate multilingual evaluation data across 13 languages~\cite{zhao2025makieval}. Although LLMs typically perform worse on low-resource languages, raising concerns about synthetic benchmark quality, manual analysis of 10,000 generated instructions in 13 Indic languages found over 99.7\% to be of high or moderate quality~\cite{chitale2025updesh}.

In our setting, automated generation was required for scale. To reduce associated risks, we applied structured filtering and answerability scoring (§\ref{sec:QAtask_methods}), discarding items below predefined thresholds. While not a substitute for native-speaker validation, this procedure provides a systematic safeguard against noise and incoherence.

\section{Benchmark Design and Evaluation Methods}
\label{sec:benchmark_design}
In this section, we describe the design of \textbf{GaelEval} and our evaluation methods. Unlike translation-based frameworks such as FLORES-200, GaelEval integrates an expert-designed morphosyntactic MCQA task with culturally grounded Gaelic-source texts for translation and Q\&A evaluation.

\subsection{Tasks}

\subsubsection{Linguistic Competence}
\label{sec:ling_task}
We define \textit{linguistic competence} as the ability to select grammatically and idiomatically appropriate forms in controlled morphosyntactic contexts. The 120-item MCQ set was designed by a Gaelic domain expert, who used a recent grammar \cite{lamb2024} to identify grammatical edge cases (e.g.~long-distance relativisation) and constructions resistant to literal translation from English.\footnote{Our approach is inspired by an unpublished Irish-language MCQA study by Joseph McInerney.} Items span 16 grammatical categories (Table~\ref{tab:mcq_distribution}), with nominal morphology the largest (17 items; 14.6\%). The design prioritised breadth across grammatical domains (e.g.\ case marking, agreement, idiomatic conventions) over depth within individual micro-phenomena (e.g.\ the feminine singular genitive), while keeping the task manageable for human participants. The MCQs were not publicly available prior to evaluation, ensuring zero-shot conditions.

\begin{table}[t]
\small
\centering
\begin{tabular}{lr}
\toprule
\textbf{Category} & \textbf{N} \\
\midrule
Nominal morphology & 17 \\
Adjectives & 11 \\
Verbal noun cores & 12 \\
Formulaic expressions & 10 \\
Questions and tags & 10 \\
Prepositions & 9 \\
Pronouns and anaphor resolution & 9 \\
Tense Aspect Modality (TAM) system & 7 \\
Impersonals and passives & 7 \\
Adverbials & 5 \\
Conjunctions and particles & 5 \\
Relative clauses & 5 \\
Clefts and focussing expressions & 4 \\
Colours & 3 \\
Determiners & 3 \\
Numerals & 3 \\
\midrule
\textbf{Total} & \textbf{120} \\
\bottomrule
\end{tabular}
\caption{Distribution of MCQs across principal grammatical categories.}
\label{tab:mcq_distribution}
\end{table}

Each question contained a single gap and was designed to have one unambiguous correct answer. For example, the following question asks for the feminine singular basic definite form of the noun \textit{fuil} `blood', where b. is the correct answer.

\begin{quote}
\begin{verbatim}
Chunnaic Màiri ___. 
('Mary saw ___.')
a. am fuil
b. an fhuil
c. am fhuil
d. na fala 
\end{verbatim}
\end{quote}

Distractors were constructed to be linguistically plausible and, as much as possible, attested in contemporary usage. This ensured that a successful choice required grammatical discrimination rather than mere lexical recognition. The placement of the correct answer was varied during preparation, but the task was issued in a fixed order across all iterations to ensure consistency for humans and models.  

Human participants were recruited via convenience sampling on social media (Facebook and LinkedIn), yielding a voluntary, non-representative sample of Gaelic speakers ($N = 35$; $n = 30$ after filtering). The task was administered online via Qualtrics. Following informed consent and instructions in English, for accessibility, responses were collected anonymously along with self-reported proficiency and age of acquisition (“Onset”). Proficiency levels were: Near-/Native, Advanced (fluent in most contexts), Upper intermediate (comfortable in most discussions), and Intermediate (everyday conversational ability). 

For benchmarking, we combined the Near-/Native and Advanced groups as `Fluent' to approximate stable adult competence. While native speakers typically acquire Gaelic in childhood, advanced and near-native speakers often receive formal instruction. This may increase familiarity with the prescriptive grammatical forms targeted in the MCQA.

The distribution of participants by onset and proficiency is shown in Table~\ref{tab:participant_distribution}. Approximately one third (12/35) identified as Near-/Native and reported acquiring Gaelic before age five, consistent with socialisation in a Gaelic-speaking home. We refer to this group as `native' in Table \ref{tab:human_proficiency_results}. 

Mean task duration was 45.7 minutes (median = 33.75; range = 12.5–165.2). Questions were presented in fixed order, and participants were required to select a single response per item. Missing responses were scored as incorrect; submissions with more than 10 missing items were excluded.

Following task administration, a few questions were flagged as potentially dialect-sensitive or admitting multiple acceptable responses. After review in light of documented Gaelic morphological variation \cite{adger,RN1167}, three items (IDs 12, 21, 48) were excluded. All reported results, therefore, are based on 120 questions versus the original 123.

\begin{table}[t]
\centering
\small
\renewcommand{\arraystretch}{1.2} 
\begin{tabular}{@{} l cccc r @{}} 
\toprule
& \multicolumn{2}{c}{\textbf{FLUENT}} & \multicolumn{2}{c}{\textbf{NON-FLUENT}} & \\
\cmidrule(lr){2-3} \cmidrule(lr){4-5}
\textbf{Onset} & \textbf{Near-/Native} & \textbf{Adv} & \textbf{Upp-Int} & \textbf{Int} & \textbf{Sum} \\
\midrule
0--4   & 12 & 0 & 0 & 0 & 12 \\
12--18 & 2  & 2 & 1 & 0 & 5  \\
18--30 & 7  & 5 & 3 & 1 & 16 \\
30+    & 2  & 0 & 0 & 0 & 2  \\
\midrule
\textbf{Total} & \textbf{23} & \textbf{7} & \textbf{4} & \textbf{1} & \textbf{35} \\
\bottomrule
\end{tabular}
\caption{Participant distribution by age of first exposure to Gaelic (Onset), self-reported proficiency level ($N=35$) and fluency grouping. `Fluent' combines near-/native and advanced.}
\label{tab:participant_distribution}
\end{table}

We input the MCQs to the models listed in Table \ref{tab:eval_models} (see §\ref{sec:LLM_models_evaluated}) with each item evaluated in a single-turn call. For each call, the prompt comprised a fixed system instruction and a user message containing the question sentence with a single blank and the full list of answer options (see §\ref{sec:ling_task_prompts}). Models were instructed to return only the text of the correct option (e.g.~\verb|b. an fhuil|), without explanation or additional punctuation, and short examples were included to enforce this format. 

Decoding parameters were left at model defaults. Responses were scored by exact string match after trimming whitespace; outputs beginning with \texttt{Error:} were logged as API failures. To mitigate transient failures and rate limits, calls were retried with exponential back-off (up to five attempts), and per-item outputs and correctness flags were stored in JSONL format. 

\subsubsection{Culturally Relevant Translation}

To construct the translation task, we collected parallel English and Gaelic transcripts of the Gaelic learning podcast \textit{An Litir Bheag} (‘The Little Letter’) from \href{https://learngaelic.scot/}{LearnGaelic.scot}. As these transcripts are professionally translated, we treat them as gold references for English–Gaelic evaluation. Produced by a fluent Gaelic speaker for intermediate and advanced learners, the episodes frequently address culturally salient topics, making them a relevant resource for assessing LLM performance on MT.

We downloaded all available episodes of \textit{An Litir Bheag} with both English and Gaelic transcripts at the time of writing (episodes 154–1076). Five episodes were excluded due to failed mp3 downloads. (We had initially intended to compile a parallel corpus including Gaelic audio to support ASR evaluation.) The final dataset comprises 918 English–Gaelic parallel transcripts.

On manual inspection, we found errors in how the podcast producers had published some of the podcasts, and so we performed filtering to ensure general data quality. First, we found that some transcripts had their identifying language reversed (i.e.~Gaelic vs English, and vice-versa), so we applied a text language identification model, OpenLID v2 \cite{burchell-etal-2023-open}, to all transcripts and manually removed those that showed switched languages.

We also identified cases where transcripts did not correspond to the correct podcast or failed to align as parallel pairs. To detect such mismatches systematically, we translated all Gaelic transcripts into English using GPT-5.2 and computed BLEU scores against the paired English versions using SacreBLEU \cite{post-2018-call}. Episodes were sorted by lowest BLEU score to identify likely mismatches. Following automated filtering and manual inspection, 10 episodes were removed, yielding a final dataset of 908 parallel transcripts.

Finally, we downloaded episode subject metadata from the LearnGaelic.scot website and aligned it with the episode transcripts. This enabled us to classify each episode according to one of the following thematic categories: Folklore, Gaelic language, History, Nature, Pastimes, People or Places.

\subsubsection{Q\&A Task on Cultural Understanding}
\label{sec:QAtask_methods}

Alongside our parallel transcripts, we also generated a MCQ set to assess the LLMs' level of Gaelic understanding and knowledge. We first instructed GPT-5.2 to rate the cultural significance of each episode's transcripts from 1-5 (system messages are detailed in the Appendix \ref{sec:scoringsysmsg}). For this task, we removed any episode rated less than 4 to filter transcripts unrelated to general knowledge of Gaelic culture (e.g.~autobiographic episodes). Following this procedure, we maintained 713 episodes out of the original 908. 


We then instructed GPT-5.2 to generate between 1 and 10 general knowledge-style Q\&A pairs per episode based on the episode transcripts in Gaelic, alongside English translations of each question and answer. 
This yielded 6,802 Q\&A pairs.

Occasionally, generated questions referred to transcript-specific details despite instructions to avoid contextual dependence, rendering them unsuitable as stand-alone general knowledge items. We therefore used GPT-5.2 to assign an ‘answerability’ score (1–5) to each Gaelic question and its English translation, where 5 denotes a fully self-contained general-knowledge item and 1 indicates contextual dependence. 
Questions scoring below 4 in either language were excluded. This filtering yielded a final set of 1,087 questions drawn from 440 unique episodes, a subset of which was manually verified by a Gaelic domain expert.

Finally, we re-input the transcripts to GPT-5.2 together with the generated questions and answers, instructing it to produce three plausible distractors per item. English translations were also generated for each distractor. 
This yielded 1,087 multiple-choice questions, each comprising one correct answer and three distractors. Answer options were randomly shuffled and labelled A–D, to prevent positional bias and ensure consistent single-letter responses from models.

\subsection{LLM Models Evaluated}
\label{sec:LLM_models_evaluated}

\begin{table}[t]
\centering
\small
\begin{tabular}{p{0.9\linewidth}}
\toprule
\textbf{Open-weight:} DeepSeek R1, GLM 4.7, GPT OSS 120B, GPT OSS 20B, Llama 4 Maverick. \\[4pt]
\textbf{Closed-weight:} Claude Haiku 4.5, Claude Opus 4.6, Gemini 2.5 Flash, Gemini 3 Flash Preview, Gemini 3 Pro Preview, GPT-4.1, GPT-4.1 Mini, GPT-4.1 Nano, GPT-4o, GPT-4o Mini, GPT-5, GPT-5 Mini, GPT-5 Nano,  GPT-5.2 \\
\bottomrule
\end{tabular}
\caption{Models evaluated in GaelEval.}
\label{tab:eval_models}
\end{table}

We evaluated 19 models (Table~\ref{tab:eval_models}) across three tasks: linguistic competence, translation and cultural understanding, spanning both open- and closed-weight systems. Models were accessed via the \mbox{OpenAI}, Anthropic, Google AI Studio and Together AI endpoints, with batch processing used to reduce costs. Responses were constrained to a predefined JSON schema (single key–value pair) to minimise explanatory output preceding the answer.

\subsection{Evaluation Metrics}
For the MCQA tasks, we report accuracy, defined as the percentage of outputs that both conformed to the required JSON schema (see Section~\ref{sec:LLM_models_evaluated}) and contained the correct answer. For translation, we report BLEU~\citep{papineni2002bleu} and \textsc{chrF}~\citep{popovic2015chrf}.

\section{Results}
\label{sec:results}

\subsection{Linguistic Competence Task}

As detailed in §\ref{sec:ling_task}, we deployed a 120-item MCQ set to assess models’ Gaelic linguistic competence. We also administered the same task to human participants ($N = 35$). Table~\ref{tab:manualqa_accuracy} reports model accuracy under two prompt conditions (Gaelic and English system messages: see §\ref{sec:ling_task_prompts}), alongside the performance of fluent speakers ($n = 30$); we exclude intermediate learners to approximate stable adult competence.

Gemini~3~Pro~Preview achieves the strongest overall results, scoring 83.3\% under Gaelic prompting and 80.0\% under English prompting. Its Gaelic-prompted score is significantly higher than the fluent-speaker mean of 78.1\% (95\% CI: 73.9\%-82.2\%; $p<0.05$). This contrasts with comparable work on Irish, where fluent speakers ($n=3$) outperformed all evaluated LLMs \cite{mcgiff2025irish}. Among \mbox{OpenAI}'s models, GPT-5 performs best (69.2\% Gaelic; 71.7\% English), while Claude~Opus~4.6 leads the Anthropic systems (59.2\% Gaelic; 50.8\% English), though it remains more than 24 percentage points below Gemini~3~Pro~Preview under Gaelic prompting.

Among open-weight systems, DeepSeek R1 (45.0\% Gaelic; 42.5\% English) and Llama 4 Maverick (45.0\% Gaelic; 40.0\% English) performed best, yet remained approximately 38 percentage points below Gemini 3 Pro Preview. Both models were only modestly above the 25\% chance baseline, consistent with results reported for other morphologically rich low-resource languages \citep{etxaniz2024latxa, mcgiff2025irish}. This gap may reflect differences in training data scale and composition between open- and closed-weight systems. A plausible factor in Google’s favour is its long-standing support for Scottish Gaelic in Google Translate (since 2016) \citep{GoogleTranslate_gaelic}, and the associated accumulation of Gaelic data.

We observe a modest aggregate advantage for Gaelic prompting. Averaged across models (excluding humans), Gaelic system messages outperform English ones by 2.4 percentage points (46.8\% vs.\ 44.4\%). Although the effect varies by model family, the overall pattern indicates a small but consistent in-language advantage. This pattern is interesting in light of \citet{myung2025blendbenchmarkllmseveryday}, who find that in-language prompting benefits medium- and high-resource languages but not typically low-resource ones. At the same time, the advantage is small and contrasts to our findings for the cultural understanding task (see §\ref{sec:cultural_understanding_results}, Table \ref{tab:openqa_acc}).

Notwithstanding the small sample size (see §\ref{sec:ling_task}), advanced speakers -- most of whom reported acquiring Gaelic between ages 18 and 30 -- outperformed native speakers on this task (82.6\% vs 70.9\%; Table~\ref{tab:human_proficiency_results}). Feasibly, some tested phenomena involve formal registers or infrequent edge cases more likely to be acquired through structured learning than everyday usage. Within our sample, self-reported fluency therefore does not fully align with prescriptive grammatical knowledge.

Within model families, we observe patterns consistent with known scaling-law behaviour. Performance improvements have been shown to follow predictable power-law trends as training compute, dataset size and parameter count increase proportionately \citep{kaplan2020scaling, hoffmann2022training}. Table~\ref{tab:manualqa_accuracy} reflects this tendency: more recent flagship models generally outperform earlier versions within the same family (e.g.\ Gemini 3 Pro: 83.3\% vs Gemini 2.5 Flash: 61.7\% in the Gaelic condition). Similar intra-family gains are observed among \mbox{OpenAI} and Anthropic models.

Smaller `Mini' and `Nano' variants -- designed to reduce parameter count and inference cost through compression and distillation -- consistently underperform their corresponding flagship models (e.g.\ GPT-5: 69.2\% vs GPT-5 Mini: 47.5\% under Gaelic prompting), reflecting the expected trade-off between efficiency and representational capacity. While recent work shows that smaller models can achieve strong performance in low-resource contexts when specifically adapted \citep{etxaniz2024latxa}, our results indicate that, without such adaptation, reduced-capacity variants perform well below large proprietary models for Gaelic.

Overall, under Gaelic prompting, Gemini 3 Pro Preview demonstrates linguistic competence in Gaelic exceeding that of fluent speakers, with GPT-5’s English-prompted performance slightly under the human baseline. Although the extent to which this task generalises to broader real-world text production remains unclear, we observe comparable performance patterns in the translation and cultural understanding tasks discussed below, supporting its construct validity.

\begin{figure*}[t]
  \centering
  \includegraphics[width=\textwidth]{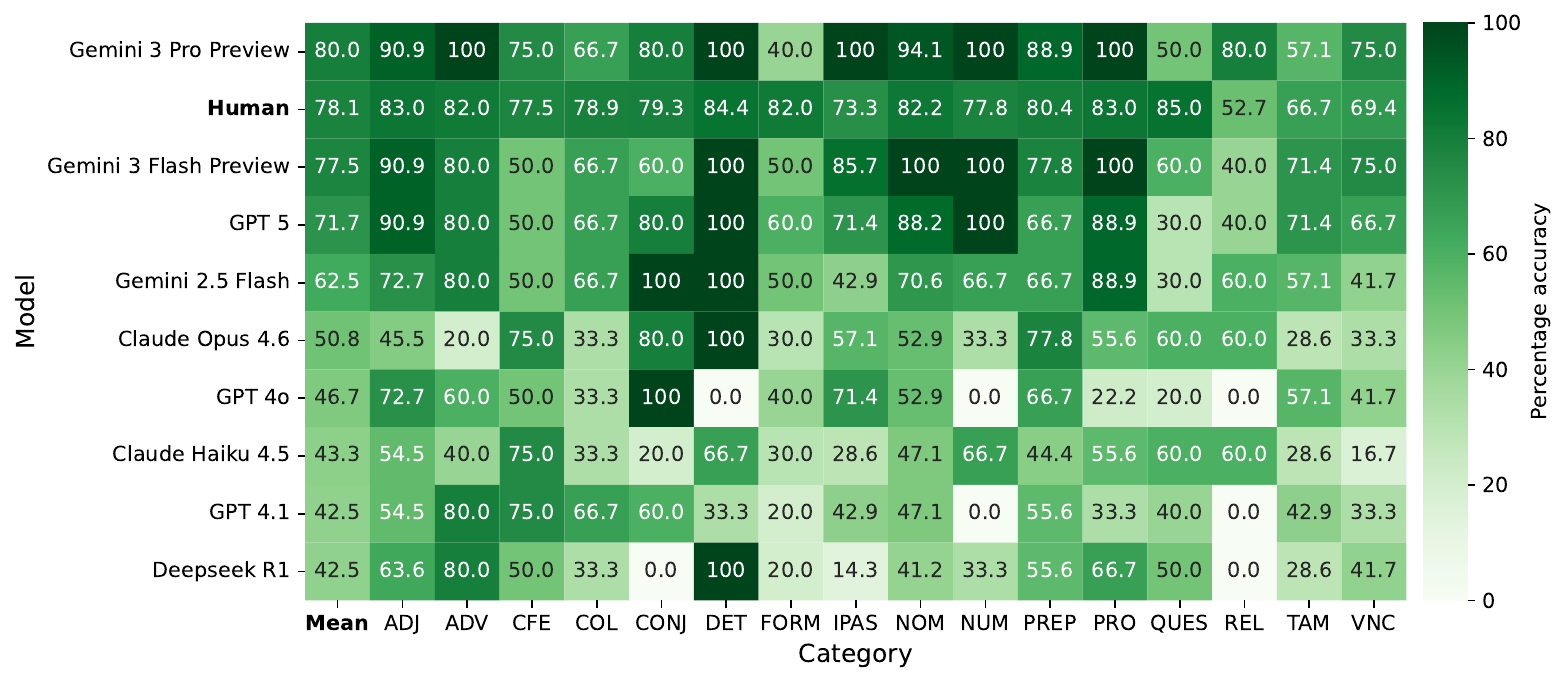}
   \caption{
  Linguistic competence accuracy by grammatical category for the human baseline and the top nine most-performant models (0--1 scale; darker = higher accuracy). ADJ = adjectives; ADV = adverbials; CFE = clefts and focussing expressions; COL = colours; CONJ = conjunctions and particles; DET = determiners; FORM = formulaic expressions; IPAS = impersonals and passives; NOM = nominal morphology; NUM = numerals; PREP = prepositions; PRO = pronouns and anaphor resolution; QUES = questions and tags; REL = relative clauses; TAM = Tense-Aspect-Modality system; VNC = verbal noun cores. Means are cross-category and so differ from those in Table \ref{tab:manualqa_accuracy}. English prompting conditions used.
  }
  \label{fig:ling_comp_heatmap}
\end{figure*}

Finally, Figure~\ref{fig:ling_comp_heatmap} disaggregates accuracy by grammatical category (cf.~Table~\ref{tab:mcq_distribution}), comparing the fluent-speaker baseline with the nine highest-performing models. Two contrasts are evident. First, models underperform humans on interactional and idiomatic material -- most clearly in questions/tags (Human = 0.85; Gemini~3~Pro = 0.50; GPT-5 = 0.30) and formulaic expressions (Human = 0.82; top models $\approx$ 0.40–0.60). Second, several models match or exceed the human mean in more structural domains, including determiners, pronouns and anaphor resolution, and relative clauses (e.g.\ REL: Human = 0.53 vs.\ Gemini~3~Pro $\approx$ 0.80). Given uneven category sizes, these differences are indicative rather than definitive. Nevertheless, if replicated with a larger, balanced MCQA, the pattern would suggest that current models handle codified morphosyntax more reliably than interactional discourse, with implications for conversational AI and CALL targeting everyday Gaelic.

\begin{table}[t]
\centering
\small
\begin{tabular}{lccc}
\toprule
Model / Group & Gaelic & English & $\Delta$ G-E \\
\midrule
Gemini 3 Pro Prev & \textbf{83.3} & \textbf{80.0} & 3.3 \\
Gemini 3 Flash Prev & 79.2 & 77.5 & 1.7 \\
Humans (Fluent) & n/a & 78.1 & -- \\
GPT-5 & 69.2 & 71.7 & -2.5 \\
Gemini 2.5 Flash & 61.7 & 62.5 & -0.8 \\
Claude Opus 4.6 & 59.2 & 50.8 & 8.3 \\
GPT-4o & 50.0 & 46.7 & 3.3 \\
Claude Haiku 4.5 & 47.5 & 43.3 & 4.2 \\
GPT-5 Mini & 47.5 & 41.7 & 5.8 \\
DeepSeek R1 & 45.0 & 42.5 & 2.5 \\
Llama 4 Maverick & 45.0 & 40.0 & 5.0 \\
GPT-5.2 & 43.3 & 40.0 & 3.3 \\
GPT-4.1 & 42.5 & 42.5 & 0.0 \\
GPT-5 Nano & 36.7 & 30.8 & 5.8 \\
GPT-4o Mini & 34.2 & 30.0 & 4.2 \\
GPT OSS 120B & 33.3 & 27.5 & 5.8 \\
GLM 4.7 & 32.5 & 26.7 & 5.8 \\
GPT-4.1 Nano & 30.0 & 31.7 & -1.7 \\
GPT OSS 20B & 29.2 & 22.5 & 6.7 \\
GPT-4.1 Mini & 24.2 & 27.5 & -3.3 \\
\midrule
\textit{Mean (models)} & 46.8 & 44.4 & 2.4 \\
\bottomrule
\end{tabular}
\caption{MCQA accuracy (\%)  under Gaelic and English prompting. $\Delta$ indicates Gaelic minus English performance. Human baseline shown for fluent speakers (n=30).}
\label{tab:manualqa_accuracy}
\end{table}

\begin{table}[t]
\centering
\small
\begin{tabular}{lrrrr}
\toprule
\textbf{Level} & \textbf{N} & \textbf{Mean} & \textbf{Max} & \textbf{Min} \\
\midrule
Near-/Native & 23 & 76.7 & 98.3 & 53.3 \\
\quad Native only & 12 & 70.9 & 85.0 & 53.3 \\
Advanced & 7 & \textbf{82.6} & 97.5 & 63.3 \\
Upper Intermediate & 4 & 73.1 & 79.2 & 67.5 \\
Intermediate & 1 & 41.7 & 41.7 & 41.7 \\
\midrule
Fluent & 30 & 78.1 & 98.3 & 53.3 \\
\bottomrule
\end{tabular}
\caption{Human accuracy (\%) on the 120-item MCQA by self-reported proficiency. `Fluent' includes near-/native and advanced speakers.}
\label{tab:human_proficiency_results}
\end{table}

\subsection{Culturally Relevant Translation}

\begin{table}[]
\centering
\resizebox{\columnwidth}{!}{%
\begin{tabular}{|l|rr|rr|}
\hline
\multirow{2}{*}{} & \multicolumn{2}{l|}{\textbf{en $\rightarrow$ gd}} & \multicolumn{2}{l|}{\textbf{gd $\rightarrow$ en}} \\ \cline{2-5} 
 & \multicolumn{1}{l|}{\textbf{BLEU}} & \multicolumn{1}{l|}{\textbf{\textsc{chrF}}} & \multicolumn{1}{l|}{\textbf{BLEU}} & \multicolumn{1}{l|}{\textbf{\textsc{chrF}}} \\ \hline
\textbf{Gemini 3 Flash Preview} & \multicolumn{1}{r|}{\textbf{71.47}} & \textbf{79.07} & \multicolumn{1}{r|}{71.75} & 77.38 \\ \hline
\textbf{Gemini 3 Pro Preview} & \multicolumn{1}{r|}{65.56} & 78.57 & \multicolumn{1}{r|}{73.41} & \textbf{78.20} \\ \hline
\textbf{Gemini 2.5 Flash} & \multicolumn{1}{r|}{65.53} & 74.72 & \multicolumn{1}{r|}{74.36} & 77.58 \\ \hline
\textbf{GPT-4.1} & \multicolumn{1}{r|}{65.41} & 74.62 & \multicolumn{1}{r|}{66.60} & 72.77 \\ \hline
\textbf{Deepseek R1} & \multicolumn{1}{r|}{62.12} & 71.84 & \multicolumn{1}{r|}{\textbf{75.28}} & 77.73 \\ \hline
\textbf{GPT-5.2} & \multicolumn{1}{r|}{61.19} & 72.16 & \multicolumn{1}{r|}{71.02} & 76.18 \\ \hline
\textbf{GPT-5} & \multicolumn{1}{r|}{56.94} & 70.27 & \multicolumn{1}{r|}{70.27} & 75.82 \\ \hline
\textbf{Claude Opus 4.6} & \multicolumn{1}{r|}{53.16} & 69.34 & \multicolumn{1}{r|}{70.37} & 77.38 \\ \hline
\textbf{GPT-4o} & \multicolumn{1}{r|}{52.39} & 67.83 & \multicolumn{1}{r|}{68.73} & 74.80 \\ \hline
\textbf{GPT-5 Mini} & \multicolumn{1}{r|}{49.19} & 60.93 & \multicolumn{1}{r|}{69.58} & 73.34 \\ \hline
\textbf{Llama 4 Maverick} & \multicolumn{1}{r|}{42.93} & 62.30 & \multicolumn{1}{r|}{73.49} & 75.74 \\ \hline
\textbf{GPT-5 Nano} & \multicolumn{1}{r|}{42.56} & 55.54 & \multicolumn{1}{r|}{61.78} & 67.41 \\ \hline
\textbf{GPT-4.1 Mini} & \multicolumn{1}{r|}{38.90} & 58.01 & \multicolumn{1}{r|}{69.09} & 75.48 \\ \hline
\textbf{Claude Haiku 4.5} & \multicolumn{1}{r|}{38.44} & 59.27 & \multicolumn{1}{r|}{72.32} & 73.92 \\ \hline
\textbf{GPT OSS 120B} & \multicolumn{1}{r|}{37.79} & 54.74 & \multicolumn{1}{r|}{56.23} & 63.01 \\ \hline
\textbf{GPT-4o Mini} & \multicolumn{1}{r|}{34.02} & 56.84 & \multicolumn{1}{r|}{69.01} & 73.29 \\ \hline
\textbf{GPT-4.1 Nano} & \multicolumn{1}{r|}{33.93} & 50.35 & \multicolumn{1}{r|}{63.31} & 69.04 \\ \hline
\textbf{GPT OSS 20B} & \multicolumn{1}{r|}{0.00} & 0.00 & \multicolumn{1}{r|}{48.21} & 53.38 \\ \hline
\textbf{GLM 4.7} & \multicolumn{1}{r|}{0.00} & 0.00 & \multicolumn{1}{r|}{0.00} & 0.00 \\ \hline
\end{tabular}%
}
\caption{BLEU and \textsc{chrF} scores for both English to Gaelic (en$\rightarrow$gd) and Gaelic to English (gd$\rightarrow$en) translation tasks.}
\label{tab:translation_results}
\end{table}


Translation metrics are reported in Table~\ref{tab:translation_results}. We observe consistent score degradation when translating into Gaelic (en$\rightarrow$gd) versus English (gd$\rightarrow$en). This corroborates prior work showing that generation in a low-resource language imposes greater representational demands than comprehension mapped back to a high-resource language such as English~\cite{goyal2022flores}.

Gaelic to English MT largely tests how well a model projects low-resource inputs into English-dominant latent spaces and then generates in English. Conversely, English to Gaelic MT demands active production of morphologically complex and culturally-grounded forms. This asymmetry is especially pronounced in GPT~5~Mini, which drops over 20 BLEU points across directions. Gemini~3~Flash~Preview is the exception, maintaining near symmetry (71.47 BLEU en$\rightarrow$gd; 71.75 BLEU gd$\rightarrow$en), suggesting a more balanced multilingual pre-training distribution.

Directional asymmetry is particularly pronounced among open-weight models. DeepSeek~R1 attains the highest BLEU score for gd$\rightarrow$en translation (75.28), marginally surpassing leading proprietary systems, but drops to 62.12 BLEU for en$\rightarrow$gd generation. This pattern suggests that while mid-tier open-weight models can parse and comprehend Gaelic, they lack the generative capacity to produce morphologically fluent output. The effect is most extreme in GPT~OSS~20B, which achieves 48.21 BLEU for gd$\rightarrow$en yet collapses to 0.00 BLEU for en$\rightarrow$gd, failing to generate valid Gaelic within the required JSON format.

We also note that GLM 4.7 was not able to form a single correctly formatted JSON response to any of our translation requests. This indicates that prompting some LLMs to process long passages of low-resource languages may degrade their ability to perform more basic tasks, such as JSON formatting.

Finally, divergence between word-level BLEU and character-level \textsc{chrF} highlight the challenges posed by Gaelic morphology. Gaelic’s rich inflectional morphology means a model may retrieve the correct lemma yet miss the surface form that BLEU requires. Accordingly, \textsc{chrF} scores are consistently higher and less variable, particularly for en$\rightarrow$gd translation, indicating difficulty with morphological realisation.

\subsection{Q\&A Task on Cultural Understanding}
\label{sec:cultural_understanding_results}

\begin{table}[t]
\centering
\small
\begin{tabular}{lccc}
\toprule
Model & Gaelic & English & $\Delta$ G-E \\
\midrule
Gemini~3~Flash~Prev & \textbf{91.35} & \textbf{91.63} & -0.28 \\
Gemini~3~Pro~Prev   & 91.17 & 90.98 & 0.19 \\
GPT-5                  & 85.28 & 88.50 & -3.22 \\
Gemini~2.5~Flash       & 79.85 & 83.44 & -3.59 \\
Claude~Opus~4.6        & 79.48 & 83.53 & -4.05 \\
GPT-5~Mini             & 71.48 & 80.04 & -8.56 \\
GPT-5.2                & 66.70 & 70.65 & -3.95 \\
GPT-4.1                & 66.24 & 74.70 & -8.46 \\
GPT-4o                 & 63.85 & 73.41 & -9.56 \\
GLM~4.7                & 63.75 & 72.40 & -8.65 \\
Llama~4~Maverick       & 59.06 & 70.01 & -10.95 \\
DeepSeek~R1            & 58.33 & 75.53 & -17.20 \\
GPT-5~Nano             & 54.37 & 70.29 & -15.92 \\
GPT~OSS~120B           & 52.81 & 70.56 & -17.75 \\
GPT-4.1~Mini           & 47.93 & 65.59 & -17.66 \\
GPT-4o~Mini            & 43.05 & 65.04 & -21.99 \\
GPT~OSS~20B            & 42.41 & 57.04 & -14.63 \\
Claude~Haiku~4.5       & 40.29 & 47.84 & -7.55 \\
GPT-4.1~Nano           & 34.87 & 51.61 & -16.74 \\
\midrule
\textit{Mean} & 63.75 & 73.30 & -9.55 \\
\bottomrule
\end{tabular}
\caption{Accuracy (\%) on the cultural knowledge Q\&A task under Gaelic and English prompting. $\Delta$ indicates Gaelic minus English performance.}
\label{tab:openqa_acc}
\end{table}

Table~\ref{tab:openqa_acc} reports accuracy on the Gaelic- and English-prompted versions of the cultural Q\&A task. Gaelic-prompted performance ranges from 34.87\% (GPT-4.1~Nano) to 91.35\% (Gemini~3~Flash~Preview), indicating substantial variation in cultural knowledge and reasoning. The strongest models -- Gemini~3~Flash~Preview, Gemini~3~Pro~Preview, and GPT-5 -- perform consistently well across both languages, with less than a 4-point gap between conditions.

In contrast to the linguistic competence task (see Table~\ref{tab:manualqa_accuracy}), most models perform worse under Gaelic prompting (mean $\Delta$ = -9.55), with the disparity widening among weaker systems. For example, GPT~OSS~20B declines from 57.04\% (English) to 42.41\% (Gaelic), while GPT-4.1~Nano drops from 51.61\% to 34.87\%, approaching chance. This pattern aligns with prior findings that LLM representations for low-resource languages are weaker than for English~\cite{goyal2022flores,romanou2024include,singh2025global}.

Comparing these results with the hand-curated Linguistic Competence task reveals strong rank correlation across benchmarks. The top three models -- Gemini~3~Flash~Preview, Gemini~3~Pro~Preview, and GPT-5 -- retain the same ordering, and mid-tier open-weight systems (e.g., DeepSeek~R1, Llama~4~Maverick) and smaller distilled variants exhibit similar relative positions.

However, absolute scores are inflated on the synthetically generated cultural QA task. The strongest models exceed 90\% accuracy, whereas the linguistic competence MCQ peaks at 83.3\%. This suggests that LLM-generated benchmarks may create easier evaluation conditions while nevertheless preserving relative performance rankings.

For low-resource languages lacking curated datasets, this result is encouraging: frontier models can generate synthetic cultural benchmarks from relevant authentic text, enabling scalable comparative evaluation. Although such datasets are not reliable as absolute measures of capability -- scores may over-estimate fluency -- they provide a low-cost and practical heuristic when creating a manual dataset would be prohibitive.

\section{Conclusion}
\label{sec:conclusion}
This paper introduces \textbf{GaelEval}, the first multi-dimensional benchmark for Scottish Gaelic, combining expert-authored morphosyntactic MCQA, culturally grounded translation and large-scale cultural knowledge evaluation. Across 19 contemporary LLMs, we observe variation in Gaelic competence, with proprietary systems outperforming open-weight models by a large margin. Notably, Gemini 3 Pro Preview exceeds the fluent-speaker baseline on the linguistic competence task, while translation and cultural Q\&A results reveal directional asymmetries and consistent performance gaps between English- and Gaelic-prompted conditions.

Category-level analysis on the linguistic competence task suggests that current models handle codified morphosyntactic structure more reliably than interactional or formulaic usage. At the same time, rank correlation across the three tasks indicates that synthetic, LLM-generated cultural benchmarks can provide reliable evaluation signals, even if absolute scores may be inflated.

For minority languages lacking established evaluation infrastructure, GaelEval demonstrates that rigorous, language-specific benchmarking is both feasible and necessary. Future work will expand human validation, balance per-category item counts and incorporate further assessment of fluency and sociolinguistic appropriateness. Robust evaluation remains essential if `shadow' LLM competence in low-resource languages is to be understood, trusted and responsibly deployed. 

\section*{Data and Code Availability}
To preserve the integrity and reusability of GaelEval as a zero-shot evaluation benchmark, the underlying dataset is not being released at this time. Public release likely would expose the data to web scraping bots, leading to its inclusion in future model pre-training corpora and compromising the benchmark’s effectiveness. To support ongoing evaluation without exposing the data, a Gaelic LLM leaderboard is planned at \url{https://eist.ac.uk}. All code for prompt construction, API interaction, and evaluation will be released in the following public repository upon publication, ensuring transparency and reproducibility: \url{https://github.com/Peter-Devine/gaeleval}.

\section*{Ethical Considerations}
Institutional ethical approval for this research was sought on 27 January 2026 and granted on 9 February 2026 by the Ethics Officer of the first author's host institution. The study involved minimal risk to participants and did not collect personally identifiable information beyond self-reported proficiency and age of acquisition.

Although no substantial personal or social risks are associated with this work, we acknowledge the environmental costs associated with large-scale model inference. While the evaluation tasks reported here required limited compute, the broader ecological impact of LLM development remains an important consideration.

\section*{Limitations}

For the linguistic task, the human sample is small and non-representative (35 participants; 30 after filtering) and the MCQA is relatively short \cite[c.f.][]{mcgiff2025irish} and unbalanced across categories. GPT-5.2 was used for data generation and filtering and was also evaluated, which may inflate its cultural Q\&A performance due to self-preference bias~\cite{wataoka2024self,xu2025deconstructing}. Moreover, the cultural Q\&A benchmark is entirely LLM-generated, although a subset was reviewed by a Gaelic domain expert.

Finally, the evaluation focuses on linguistic and cultural accuracy rather than general reasoning or mathematical ability, as commonly assessed in English-language benchmarks. Models may exhibit different reasoning performance when operating in a low-resource language setting.

\section*{Acknowledgements}
This work was carried out by members of the CLARIN Knowledge Centre for Digital Resources for the Languages in Ireland and Britain (DR-LIB) as part of the project 'Unlocking AI for the Languages in Britain and Ireland' project, funded by EPSRC (project number UKRI3181).

\section*{References}\label{sec:reference}
\bibliographystyle{lrec2026-natbib}
\bibliography{scholarship.bib}

@inproceedings{burchell-etal-2023-open,
    title = "An Open Dataset and Model for Language Identification",
    author = "Burchell, Laurie  and
      Birch, Alexandra  and
      Bogoychev, Nikolay  and
      Heafield, Kenneth",
    editor = "Rogers, Anna  and
      Boyd-Graber, Jordan  and
      Okazaki, Naoaki",
    booktitle = "Proceedings of the 61st Annual Meeting of the Association for Computational Linguistics (Volume 2: Short Papers)",
    month = jul,
    year = "2023",
    address = "Toronto, Canada",
    publisher = "Association for Computational Linguistics",
    url = "https://aclanthology.org/2023.acl-short.75",
    doi = "10.18653/v1/2023.acl-short.75",
    pages = "865--879",
}

@inproceedings{post-2018-call,
  title = "A Call for Clarity in Reporting {BLEU} Scores",
  author = "Post, Matt",
  booktitle = "Proceedings of the Third Conference on Machine Translation: Research Papers",
  month = oct,
  year = "2018",
  address = "Belgium, Brussels",
  publisher = "Association for Computational Linguistics",
  url = "https://www.aclweb.org/anthology/W18-6319",
  pages = "186--191",
}

@article{wataoka2024self,
  title={Self-preference bias in {LLM}-as-a-judge},
  author={Wataoka, Koki and Takahashi, Tsubasa and Ri, Ryokan},
  journal={arXiv preprint arXiv:2410.21819},
  year={2024},
  url={https://arxiv.org/abs/2410.21819}
}

@inproceedings{papineni2002bleu,
  title={{BLEU}: a method for automatic evaluation of machine translation},
  author={Papineni, Kishore and Roukos, Salim and Ward, Todd and Zhu, Wei-Jing},
  booktitle={Proceedings of the 40th annual meeting of the Association for Computational Linguistics},
  pages={311--318},
  year={2002},
  url={https://aclanthology.org/P02-1040.Pdf}
}

@inproceedings{popovic2015chrf,
  title={{chrF}: character n-gram F-score for automatic MT evaluation},
  author={Popovi{\'c}, Maja},
  booktitle={Proceedings of the tenth workshop on statistical machine translation},
  pages={392--395},
  year={2015},
  url={https://aclanthology.org/W15-3049.pdf}
}

@misc{GoogleTranslate_gaelic,
   author = {BBC},
   title = {Google Translate introduces 13 new languages including Scots Gaelic and Sindhi},
   month = {18 February 2016},
   url = {https://www.bbc.co.uk/news/newsbeat-35602377},
   year = {2016},
   type = {Newspaper Article}
}

@inproceedings{Wendler2024,
  title={Do llamas work in english? on the latent language of multilingual transformers},
  author={Wendler, Chris and Veselovsky, Veniamin and Monea, Giovanni and West, Robert},
  booktitle={Proceedings of the 62nd Annual Meeting of the Association for Computational Linguistics (Volume 1: Long Papers)},
  pages={15366--15394},
  year={2024},
  url = {https://aclanthology.org/2024.acl-long.820/}
}

@article{tao2024cultural,
  title={Cultural bias and cultural alignment of large language models},
  author={Tao, Yan and Viberg, Olga and Baker, Ryan S and Kizilcec, Ren{\'e} F},
  journal={PNAS Nexus},
  volume={3},
  number={9},
  pages={1--9},
  year={2024},
  publisher={Oxford University Press US},
  url={https://doi.org/10.1093/pnasnexus/pgae346}
}

@book{lamb2024,
   author = {Lamb, William},
   title = {Scottish Gaelic: A Comprehensive Grammar},
   publisher = {Routledge},
   address = {Oxon},
   year = {2024},
   type = {Book}
}

@inproceedings{joshi-etal-2020-state,
    title = "The State and Fate of Linguistic Diversity and Inclusion in the {NLP} World",
    author = "Joshi, Pratik  and
      Santy, Sebastin  and
      Budhiraja, Amar  and
      Bali, Kalika  and
      Choudhury, Monojit",
    editor = "Jurafsky, Dan  and
      Chai, Joyce  and
      Schluter, Natalie  and
      Tetreault, Joel",
    booktitle = "Proceedings of the 58th Annual Meeting of the Association for Computational Linguistics",
    month = jul,
    year = "2020",
    address = "Online",
    publisher = "Association for Computational Linguistics",
    url = "https://aclanthology.org/2020.acl-main.560/",
    doi = "10.18653/v1/2020.acl-main.560",
    pages = "6282--6293"
}

@article{RN1167,
   author = {Iosad, Pavel and Lamb, William},
   title = {Dialect variation in Scottish Gaelic nominal morphology: A quantitative study},
   journal = {Glossa},
   volume = {5},
   number = {1},
   pages = {1–31},
   DOI = {https://doi.org/10.5334/gjgl.1023},
   year = {2020},
   type = {Journal Article}
}

@incollection{adger,
  author    = {Adger, David},
  title     = {Gaelic morphology},
  booktitle = {The Edinburgh Companion to the Gaelic Language},
  editor    = {Watson, Moray and Macleod, Michelle},
  publisher = {Edinburgh University Press},
  address   = {Edinburgh},
  pages     = {283--303},
  year      = {2010}
}

@article{goyal2022flores,
  title={The {FLORES}-101 Evaluation Benchmark for Low-Resource and Multilingual Machine Translation},
  author={Goyal, Naman and Gao, Cynthia and Chaudhary, Vishrav and Chen, Peng-Jen and Wenzek, Guillaume and Ju, Da and Krishnan, Sanjana and Ranzato, Marc’Aurelio and Guzm{\'a}n, Francisco and Fan, Angela},
  journal={Transactions of the Association for Computational Linguistics},
  volume={10},
  pages={522--538},
  year={2022},
  publisher={MIT Press-Journals},
  url={https://doi.org/10.1162/tacl_a_00474}
}

@article{xia2025evaluating,
  title={Evaluating Modern Large Language Models on Low-Resource and Morphologically Rich Languages: A Cross-Lingual Benchmark Across {C}antonese, {J}apanese, and {T}urkish},
  author={Xia, Chengxuan and Wu, Qianye and Guan, Hongbin and Tian, Sixuan and Hao, Yilun and Wu, Xiaoyu},
  journal={arXiv preprint arXiv:2511.10664},
  year={2025},
  url={https://arxiv.org/abs/2511.10664}
}

@article{romanou2024include,
  title={{INCLUDE}: Evaluating multilingual language understanding with regional knowledge},
  author={Romanou, Angelika and Foroutan, Negar and Sotnikova, Anna and Chen, Zeming and Nelaturu, Sree Harsha and Singh, Shivalika and Maheshwary, Rishabh and Altomare, Micol and Haggag, Mohamed A and Amayuelas, Alfonso and others},
  journal={arXiv preprint arXiv:2411.19799},
  year={2024},
  url={https://doi.org/10.48550/arXiv.2411.19799}
}

@inproceedings{singh2025global,
  title={{Global MMLU}: Understanding and addressing cultural and linguistic biases in multilingual evaluation},
  author={Singh, Shivalika and Romanou, Angelika and Fourrier, Cl{\'e}mentine and Adelani, David Ifeoluwa and Ngui, Jian Gang and Vila-Suero, Daniel and Limkonchotiwat, Peerat and Marchisio, Kelly and Leong, Wei Qi and Susanto, Yosephine and others},
  booktitle={Proceedings of the 63rd Annual Meeting of the Association for Computational Linguistics (Volume 1: Long Papers)},
  pages={18761--18799},
  year={2025},
  url={https://aclanthology.org/2025.acl-long.919/}
}

@article{hendrycks2020measuring,
  title={Measuring massive multitask language understanding},
  author={Hendrycks, Dan and Burns, Collin and Basart, Steven and Zou, Andy and Mazeika, Mantas and Song, Dawn and Steinhardt, Jacob},
  journal={arXiv preprint arXiv:2009.03300},
  url={https://arxiv.org/abs/2009.03300}, 
  year={2020}
}

@article{nllb-24,
    author="{NLLB Team} and Costa-juss{\`a}, Marta R. and Cross, James and {\c{C}}elebi, Onur and Elbayad, Maha and Heafield, Kenneth and Heffernan, Kevin and Kalbassi, Elahe and Lam, Janice and Licht, Daniel and Maillard, Jean and Sun, Anna and Wang, Skyler and Wenzek, Guillaume and Youngblood, Al and Akula, Bapi and Barrault, Loic and Gonzalez, Gabriel Mejia and Hansanti, Prangthip and Hoffman, John and Jarrett, Semarley and Sadagopan, Kaushik Ram and Rowe, Dirk and Spruit, Shannon and Tran, Chau and Andrews, Pierre and Ayan, Necip Fazil and Bhosale, Shruti and Edunov, Sergey and Fan, Angela and Gao, Cynthia and Goswami, Vedanuj and Guzm{\'a}n, Francisco and Koehn, Philipp and Mourachko, Alexandre and Ropers, Christophe and Saleem, Safiyyah and Schwenk, Holger and Wang, Jeff",
    title="Scaling neural machine translation to 200 languages",
    journal="Nature",
    year="2024",
    volume="630",
    number="8018",
    pages="841--846",
    issn="1476-4687",
    doi="10.1038/s41586-024-07335-x",
    url="https://doi.org/10.1038/s41586-024-07335-x"
}

@inproceedings{etxaniz2024latxa,
  title={Latxa: An open language model and evaluation suite for {B}asque},
  author={Etxaniz, Julen and Sainz, Oscar and Miguel, Naiara and Aldabe, Itziar and Rigau, German and Agirre, Eneko and Ormazabal, Aitor and Artetxe, Mikel and Soroa, Aitor},
  booktitle={Proceedings of the 62nd Annual Meeting of the Association for Computational Linguistics (Volume 1: Long Papers)},
  pages={14952--14972},
  year={2024},
  url={https://aclanthology.org/2024.acl-long.799.pdf}
}

@article{hoffmann2022training,
  title={Training compute-optimal large language models},
  author={Hoffmann, Jordan and Borgeaud, Sebastian and Mensch, Arthur and Buchatskaya, Elena and Cai, Trevor and Rutherford, Eliza and Casas, DDL and Hendricks, Lisa Anne and Welbl, Johannes and Clark, Aidan and others},
  journal={arXiv preprint arXiv:2203.15556},
  volume={10},
  year={2022},
  url={https://proceedings.neurips.cc/paper/2022/file/c1e2faff6f588870935f114ebe04a3e5-Paper-Conference.pdf}
}

@inproceedings{ruder2023xtreme,
    title = "{XTREME}-{UP}: A User-Centric Scarce-Data Benchmark for Under-Represented Languages",
    author = "Ruder, Sebastian  and
      Clark, Jonathan H.  and
      Gutkin, Alexander  and
      Kale, Mihir  and
      Ma, Min  and
      Nicosia, Massimo  and
      Rijhwani, Shruti  and
      Riley, Parker  and
      Sarr, Jean-Michel A-  and
      Wang, Xinyi  and
      Wieting, John  and
      Gupta, Nitish  and
      Katanova, Anna  and
      Kirov, Christo  and
      Dickinson, Dana L.  and
      Roark, Brian  and
      Samanta, Bidisha  and
      Tao, Connie  and
      Adelani, David I.  and
      Axelrod, Vera  and
      Caswell, Isaac  and
      Cherry, Colin  and
      Garrette, Dan  and
      Ingle, Reeve  and
      Johnson, Melvin  and
      Panteleev, Dmitry  and
      Talukdar, Partha",
    editor = "Bouamor, Houda  and
      Pino, Juan  and
      Bali, Kalika",
    booktitle = "Findings of the Association for Computational Linguistics: EMNLP 2023",
    month = dec,
    year = "2023",
    address = "Singapore",
    publisher = "Association for Computational Linguistics",
    url = "https://aclanthology.org/2023.findings-emnlp.125/",
    doi = "10.18653/v1/2023.findings-emnlp.125",
    pages = "1856--1884",
    }

@inproceedings{singh2024indicgenbench,
    title = "{I}ndic{G}en{B}ench: A Multilingual Benchmark to Evaluate Generation Capabilities of {LLM}s on {I}ndic Languages",
    author = "Singh, Harman  and
      Gupta, Nitish  and
      Bharadwaj, Shikhar  and
      Tewari, Dinesh  and
      Talukdar, Partha",
    editor = "Ku, Lun-Wei  and
      Martins, Andre  and
      Srikumar, Vivek",
    booktitle = "Proceedings of the 62nd Annual Meeting of the Association for Computational Linguistics (Volume 1: Long Papers)",
    month = aug,
    year = "2024",
    address = "Bangkok, Thailand",
    publisher = "Association for Computational Linguistics",
    url = "https://aclanthology.org/2024.acl-long.595/",
    doi = "10.18653/v1/2024.acl-long.595",
    pages = "11047--11073"
}

@inproceedings{ogundepo2023afriqa,
  title={Afriqa: Cross-lingual open-retrieval question answering for african languages},
  author={Ogundepo, Odunayo and Gwadabe, Tajuddeen R and Rivera, Clara E and Clark, Jonathan H and Ruder, Sebastian and Adelani, David Ifeoluwa and Dossou, Bonaventure FP and Diop, Abdou Aziz and Sikasote, Claytone and Hacheme, Gilles and others},
  booktitle={Findings of the Association for Computational Linguistics: EMNLP 2023},
  pages={14957--14972},
  year={2023},
  url={https://doi.org/10.18653/v1/2023.findings-emnlp.997}
}

@article{toraman2026turkbench,
  title={{TurkBench}: A Benchmark for Evaluating {T}urkish Large Language Models},
  author={Toraman, {\c{C}}a{\u{g}}r{\i} and Sever, Ahmet Kaan and Cengiz, Ayse Aysu and Arslan, Elif Ecem and Sevin{\c{c}}, G{\"o}rkem and Birdal, Mete Mert and G{\"u}ldemir, Yusuf Faruk and Kanburo{\u{g}}lu, Ali Bu{\u{g}}ra and Feleko{\u{g}}lu, Sezen and G{\"u}rlek, Osman and others},
  journal={arXiv preprint arXiv:2601.07020},
  year={2026},
  url={https://arxiv.org/abs/2601.07020}
}

@inproceedings{azime2025proverbeval,
    title = "{P}roverb{E}val: Exploring {LLM} Evaluation Challenges for Low-resource Language Understanding",
    author = "Azime, Israel Abebe  and
      Tonja, Atnafu Lambebo  and
      Belay, Tadesse Destaw  and
      Chanie, Yonas  and
      Balcha, Bontu Fufa  and
      Abadi, Negasi Haile  and
      Ademtew, Henok Biadglign  and
      Nerea, Mulubrhan Abebe  and
      Yadeta, Debela Desalegn  and
      Geremew, Derartu Dagne  and
      Tesfu, Assefa Atsbiha  and
      Slusallek, Philipp  and
      Solorio, Thamar  and
      Klakow, Dietrich",
    editor = "Chiruzzo, Luis  and
      Ritter, Alan  and
      Wang, Lu",
    booktitle = "Findings of the Association for Computational Linguistics: NAACL 2025",
    month = apr,
    year = "2025",
    address = "Albuquerque, New Mexico",
    publisher = "Association for Computational Linguistics",
    url = "https://aclanthology.org/2025.findings-naacl.350/",
    doi = "10.18653/v1/2025.findings-naacl.350",
    pages = "6265--6281",
    ISBN = "979-8-89176-195-7"
}

@inproceedings{perez2023discovering,
  title={Discovering language model behaviors with model-written evaluations},
  author={Perez, Ethan and Ringer, Sam and Lukosiute, Kamile and Nguyen, Karina and Chen, Edwin and Heiner, Scott and Pettit, Craig and Olsson, Catherine and Kundu, Sandipan and Kadavath, Saurav and others},
  booktitle={Findings of the association for computational linguistics: ACL 2023},
  pages={13387--13434},
  year={2023},
  url={https://doi.org/10.18653/v1/2023.findings-acl.847}
}

@misc{assadi2025humemeasuringhumanmodelperformance,
      title={HUME: Measuring the Human-Model Performance Gap in Text Embedding Tasks}, 
      author={Adnan El Assadi and Isaac Chung and Roman Solomatin and Niklas Muennighoff and Kenneth Enevoldsen},
      year={2025},
      eprint={2510.10062},
      archivePrefix={arXiv},
      primaryClass={cs.CL},
      url={https://arxiv.org/abs/2510.10062}, 
}

@article{mcgiff2025irish,
  title={{Irish-BLiMP:} A Linguistic Benchmark for Evaluating Human and Language Model Performance in a Low-Resource Setting},
  author={McGiff, Josh and Tran, Khanh-Tung and Mulcahy, William and Luin{\'\i}n, D{\'a}ibhidh {\'O} and Dalzell, Jake and Bhroin, R{\'o}is{\'\i}n N{\'\i} and Burke, Adam and O'Sullivan, Barry and Nguyen, Hoang D and Nikolov, Nikola S},
  journal={arXiv preprint arXiv:2510.20957},
  year={2025},
  url={https://doi.org/10.48550/arXiv.2510.20957}
}

@article{anwar2026mcqs,
  title={{MCQs} Generation With Large Language Models: A Survey of Methodologies, Evolution, and Open Research Issues},
  author={Anwar, Muhammad Raheel and Khalid, Shah and Alshahrani, Saied and Bilal, Hafiz Syed Muhammad and Aldawsari, Mohammed},
  journal={IEEE Access},
  volume={14},
  pages={10991--11018},
  year={2026},
  publisher={IEEE},
  url={https://doi.org/10.1109/ACCESS.2026.3652006}
}

@misc{myung2025blendbenchmarkllmseveryday,
      title={{BLEnD}: A Benchmark for LLMs on Everyday Knowledge in Diverse Cultures and Languages}, 
      author={Junho Myung and Nayeon Lee and Yi Zhou and Jiho Jin and Rifki Afina Putri and Dimosthenis Antypas and Hsuvas Borkakoty and Eunsu Kim and Carla Perez-Almendros and Abinew Ali Ayele and Víctor Gutiérrez-Basulto and Yazmín Ibáñez-García and Hwaran Lee and Shamsuddeen Hassan Muhammad and Kiwoong Park and Anar Sabuhi Rzayev and Nina White and Seid Muhie Yimam and Mohammad Taher Pilehvar and Nedjma Ousidhoum and Jose Camacho-Collados and Alice Oh},
      year={2025},
      eprint={2406.09948},
      archivePrefix={arXiv},
      primaryClass={cs.CL},
      url={https://arxiv.org/abs/2406.09948}, 
}

@article{kaplan2020scaling,
  title={Scaling laws for neural language models},
  author={Kaplan, Jared and McCandlish, Sam and Henighan, Tom and Brown, Tom B and Chess, Benjamin and Child, Rewon and Gray, Scott and Radford, Alec and Wu, Jeffrey and Amodei, Dario},
  journal={arXiv preprint arXiv:2001.08361},
  year={2020},
  url={https://arxiv.org/pdf/2001.08361/1000}
}

@inproceedings{papadimitriou-etal-2023-multilingual,
    title = "Multilingual {BERT} has an accent: Evaluating {E}nglish influences on fluency in multilingual models",
    author = "Papadimitriou, Isabel  and
      Lopez, Kezia  and
      Jurafsky, Dan",
    editor = "Vlachos, Andreas  and
      Augenstein, Isabelle",
    booktitle = "Findings of the Association for Computational Linguistics: EACL 2023",
    month = may,
    year = "2023",
    address = "Dubrovnik, Croatia",
    publisher = "Association for Computational Linguistics",
    url = "https://aclanthology.org/2023.findings-eacl.89/",
    doi = "10.18653/v1/2023.findings-eacl.89",
    pages = "1194--1200"
}

@inproceedings{adelani2024sib,
  title={{SIB-200}: A simple, inclusive, and big evaluation dataset for topic classification in 200+ languages and dialects},
  author={Adelani, David Ifeoluwa and Liu, Hannah and Shen, Xiaoyu and Vassilyev, Nikita and Alabi, Jesujoba and Mao, Yanke and Gao, Haonan and Lee, En-Shiun Annie},
  booktitle={Proceedings of the 18th Conference of the European Chapter of the Association for Computational Linguistics (Volume 1: Long Papers)},
  pages={226--245},
  year={2024},
  url={https://aclanthology.org/2024.eacl-long.14/}
}

@inproceedings{nguyen2023extracting,
  title={Extracting cultural commonsense knowledge at scale},
  author={Nguyen, Tuan-Phong and Razniewski, Simon and Varde, Aparna and Weikum, Gerhard},
  booktitle={Proceedings of the ACM web conference 2023},
  pages={1907--1917},
  year={2023},
  url={https://doi.org/10.1145/3543507.3583535}
}

@inproceedings{shi2024culturebank,
  title={{CultureBank:} An online community-driven knowledge base towards culturally aware language technologies},
  author={Shi, Weiyan and Li, Ryan and Zhang, Yutong and Ziems, Caleb and Yu, Sunny and Horesh, Raya and De Paula, Rog{\'e}rio Abreu and Yang, Diyi},
  booktitle={Findings of the Association for Computational Linguistics: EMNLP 2024},
  pages={4996--5025},
  year={2024},
  url={https://doi.org/10.18653/v1/2024.findings-emnlp.288}
}

@inproceedings{wang2024craft,
  title={{CRAFT}: Extracting and tuning cultural instructions from the wild},
  author={Wang, Bin and Lin, Geyu and Liu, Zhengyuan and Wei, Chengwei and Chen, Nancy},
  booktitle={Proceedings of the 2nd Workshop on Cross-Cultural Considerations in NLP},
  pages={42--47},
  year={2024},
  url={https://aclanthology.org/2024.c3nlp-1.4/}
}

@article{fung2024no,
  title={No culture left behind: Massively multi-cultural knowledge acquisition \& {LM} benchmarking on 1000+ sub-country regions and 2000+ ethnolinguistic groups},
  author={Fung, Yi R and Sun, Chenkai and Doo, Jae and Zhao, Ruining and Ji, Heng},
  journal={arxiv},
  url={https://arxiv.org/pdf/2402.09369},
  year={2024}
}

@article{zhao2025makieval,
  title={{MAKIEval}: A Multilingual Automatic {W}iKidata-based Framework for Cultural Awareness Evaluation for {LLMs}},
  author={Zhao, Raoyuan and Chen, Beiduo and Plank, Barbara and Hedderich, Michael A},
  journal={arXiv preprint arXiv:2505.21693},
  year={2025},
  url={https://aclanthology.org/anthology-files/pdf/findings/2025.findings-emnlp.1256.pdf}
}

@article{chitale2025updesh,
  title={{UPDESH}: Synthesizing Grounded Instruction Tuning Data for 13 Indic},
  author={Chitale, Pranjal A and Gumma, Varun and Ahuja, Sanchit and Kodali, Prashant and Uppadhyay, Manan and Sudharsan, Deepthi and Sitaram, Sunayana},
  journal={arXiv preprint arXiv:2509.21294},
  year={2025},
  url={https://doi.org/10.48550/arXiv.2509.21294}
}

@misc{britllm2026,
  author        = {{BritLLM}},
  title         = {{BritLLM}: {F}reely available large language models for {UK} languages and use-cases},
  howpublished = {Wayback Machine archive of \url{https://llm.org.uk/}},
  year         = {2026},
  note         = {Archived January 21, 2026. Available at: \url{https://web.archive.org/web/20260121124627/https://llm.org.uk/}},
  urldate      = {2026-02-25}
}

@article{xu2025deconstructing,
  title={Deconstructing Self-Bias in {LLM}-generated Translation Benchmarks},
  author={Xu, Wenda and Agrawal, Sweta and Zouhar, Vil{\'e}m and Freitag, Markus and Deutsch, Daniel},
  journal={arXiv preprint arXiv:2509.26600},
  year={2025},
  url={https://arxiv.org/abs/2509.26600}
}


\newpage
\section{Appendix}

\subsection{Linguistic Competence Task: System Messages}
\label{sec:ling_task_prompts}

\paragraph{English} 
You are a knowledgeable assistant that can answer all kinds of questions. Please select the correct option. Output ONLY the letter of the correct option, without any additional explanation or punctuation.
\newline \newline
\noindent Examples:

\begin{quote}
What colour is the sky? ['A. blue', 'B. yellow', 'C. green']. Return ONLY 'A'

Which of these countries is in Africa? ['A. Germany', 'B. Mexico', 'C. Nigeria']. Return ONLY 'C'
    
\end{quote}

\paragraph{Gaelic}
Is e cuidiche fiosrachail a tha annad agus is urrainn dhut a h-uile seòrsa ceist a fhreagairt. Tagh an romhainn cheart. Na cuir a-mach ach litir na freagairte ceirte, as aonais mìneachadh no pongachadh sam bith eile.
\newline \newline
\noindent Eisimpleirean:

\begin{quote}

Dè 'n dath a th' air an iarmailt? ['A. gorm', 'B. buidhe', 'C. uaine']. Na cuir a-mach ach 'A'

Cò 'n dùthaich às na leanas a tha ann an Afraca? ['A. A' Ghearmailt', 'B. Meagsago', 'C. Nigèiria']. Na cuir a-mach ach 'C' 
\end{quote}

\subsection{Transcript Scoring System Message}
\label{sec:scoringsysmsg}

You are a Scottish Gaelic article scoring assistant.
Given a Gaelic article transcript and its English translation, give a score between 1 and 5 for the cultural relevancy of the article to Gaelic culture.
If the article contains no material that relates to Gaelic culture, then give a score of 1.
If the article is full of information on important aspects of Gaelic culture, then give a score of 5.
For articles somewhere in-between these two, then give a score most fitting the content.

\subsection{Cultural Understanding QA system message}
\label{sec:openqasysmsg}

You are a Scottish Gaelic question and answer generating assistant.
Given a Gaelic article transcript and its English translation, write between 1 and 10 question in Gaelic about the content within the article.
The questions should test the answerer's knowledge of Gaelic culture in some way, using only the article as the factual basis for the question and answer.
The answers to the questions should not be easily guessed from the question.
Only include as many questions as you are able to make out of the content of the article.
Each question should be written so that it makes total sense in isolation and can be answered by someone knowledgeable on the subject without reading the article.
Make sure the questions are self contained.
You may introduce people, things, places etc. from the article in each question if that helps make the question understandable without reading the article.
Do not refer to entities contextually - always use a persons name rather than using 'the man', for example, where possible.
Each question can be as long as you would like but should be answerable in less than 10 words.
Write the questions and answers in Gaelic and also write an English translation of each.

\subsection{Answerability System Message}
\label{sec:answerabilitysysmsg}

You are a Scottish Gaelic and English question scoring assistant.
Given a question in Gaelic and its English translation, give a score between 1-5 on the self-contained answerability of both questions.
Give a score of 5 if the question is a good general knowledge question, is self-contained, is not contextual dependent, and can be answered purely from using knowledge of Gaelic culture.
Give a score of 1 if the question is contextual dependent, refers to implicit information outside of general knowledge (e.g.~'the man' rather than 'Robert the Bruce'), or is otherwise badly written.
For questions somewhere in-between these two, then give a score most fitting the content.
Evaluate the question in both languages on an individual basis, evaluating the wording of the question purely in that language.

\subsection{Distractor System Message}
\label{sec:distractorsysmsg}

You are a Scottish Gaelic distractor generating assistant.
Given a Gaelic article transcript and its English translation, as well as a question and answer about the content within the article, write three distractors for the answer.
The distractors should be incorrect answers to the question.
The distractors should also be plausible while remaining wrong answers.
If the correct answer is written in a particular style or format, make sure the distractors also follow this style or format.
Remember that there may be multiple correct answers to the original question, so make sure that your three distractors are all completely INCORRECT answers.
Write the distractors in Gaelic and also write an English translation for each.

\end{document}